\documentclass[conference]{IEEEtran}
\IEEEoverridecommandlockouts

\usepackage{cite}
\usepackage{amsmath,amssymb,amsfonts}
\usepackage{url}
\usepackage{graphicx}
\usepackage{textcomp}
\usepackage{caption}
\usepackage{algorithmic}
\usepackage{booktabs}
\usepackage{multirow}
\usepackage[dvipsnames]{xcolor}
\usepackage{scrextend}
\usepackage{enumitem}
\newlist{todolist}{itemize}{2}
\setlist[todolist]{label=$\square$}
\usepackage{pifont}

\def\BibTeX{{\rm B\kern-.05em{\sc i\kern-.025em b}\kern-.08em
    T\kern-.1667em\lower.7ex\hbox{E}\kern-.125emX}}
\begin{document}

\title{Benchmarking Fast Domain Adaptation for Unsupervised Speech Units}

\def\x{{\mathbf x}}
\def\L{{\cal L}}

\def\dataset{{ABX-Accent}}




\author{\IEEEauthorblockN{Robin San Roman\textsuperscript{\textsection}}
\IEEEauthorblockA{\textit{CoML team, ENS, PSL university,}\\
\textit{CNRS, EHESS}\\
Paris, France}
\and
\IEEEauthorblockN{Manel Khentout\textsuperscript{\textsection}}
\IEEEauthorblockA{\textit{CoML team, ENS, PSL university,}\\
\textit{CNRS, EHESS}\\
Paris, France}
\and
\IEEEauthorblockN{Tu Anh Nguyen}
\IEEEauthorblockA{\textit{CoML team, ENS, PSL university,}\\
\textit{CNRS, EHESS}\\
Paris, France}
\and
\IEEEauthorblockN{Paul Michel}
\IEEEauthorblockA{\textit{CoML team, ENS, PSL university,}\\
\textit{CNRS, EHESS}\\
Paris, France}
\and
\IEEEauthorblockN{Yossi Adi}
\IEEEauthorblockA{\textit{Hebrew University of Jerusalem}\\
Jerusalem, Israel}
\and
\IEEEauthorblockN{Emmanuel Dupoux}
\IEEEauthorblockA{\textit{CoML team, ENS, PSL university,}\\
\textit{CNRS, EHESS}\\
Paris, France}
}

\maketitle

\begingroup\renewcommand\thefootnote{\textsection}
\footnotetext{Equal contribution}

\begin{abstract}

Representation learning has attracted great attention and managed to reach good performances as a pretraining method for downstream tasks or as a first step towards unsupervised speech modeling. Yet, little is known about how such methods deal with out-of-domain speech and how could they be adapted in a few shot to new domains. This is important especially for accented speech where one observes a long tail of accents that diverge from the standard ones. We introduce \dataset{}, a benchmark based on the AESRC dataset that features 10 different accents of English. It includes a small ($<$ 10 hours) unlabelled training set in each of the accents and adaptations of the Zero Resources Challenge ABX evaluation metrics to each of the accents. We illustrate this benchmark with a baseline model that uses adaptive domain normalization to fine tune a pretrained Contrastive Predictive Coding model on the accents. This method is first developed on LibriSpeech using a male/female split. When applied to the new benchmark, the proposed method yields a relative improvement of 23.6\% on across-speaker ABX scores on average compared to non adapted models. The data and metrics will be open sourced upon paper acceptance.

\end{abstract}

\section{Introduction}

Self-supervised speech representation learning (SSL) has recently gained a lot of traction, both as a way to pretrain models on large amounts of unlabelled data, which can be fine tuned for a variety of downstream tasks \cite{yang2021superb, mohamed2022self}, and as a building block for textless NLP systems that learn language models directly from speech \cite{lakhotia21gslm,borsos22audiolm}. The assumption in both cases is that it is relatively easier to find large amounts of unlabelled audio than it is to find labelled data. Yet, even unlabeled audio is present in skewed distributions, with a predominance of high resourced languages (e.g., English, Spanish, etc.), and within these, a predominance of a few dominant dialects or accents. It is likely that as in other areas of machine learning, self-supervised speech representation learning may suffer when the data distribution is heavily imbalanced, creating bias in favor of the dominant mode, and even more when trained models are used out-of-domain. 

Here, we focus on the problem of accented speech, which has the characteristics that it is impossible to ever consider a training set that will encompass all possible accents. Sociolinguistic studies show that language's accent vary  geographically, through time and across generations. In addition, second language learners bring in their own idiosyncratic accents. This yields a long tail of constantly renewed accents which raises problems of lack of inclusiveness and bias in the downstream applications built with such datasets. A possible way to deal with this situation is to setup systems that can adapt to each new accent with as little data as possible. 

Building on the AESRC dataset that features 10 different accents of English, and present two contributions.  (1), we setup a \dataset{}, a benchmark that evaluates how SSL models can adapt to novel accents with little data and no supervision (around 10 hours per accent, with additional speaker-level adaptation data at test time). The evaluation metric is ABX \cite{schatz}, a metric used in the zero resource speech challenge series \cite{dunbar2020, dunbar} focused on the phonetic quality of the learned representations and which has been shown to predict downstream language modeling tasks based on these representations \cite{lakhotia21gslm}.  (2) we explore some simple methods for domain adaptation to provide baseline results in our new benchmark.

\section{Related Work}

\subsection{Representation learning}
Inspired by successes in computer vision, and natural language processing, self supervised learning (SSL) applied to speech, aims at learning to extract features from unlabelled data that can be used for downstream tasks such as ASR, voice conversion, emotion recognition and other tasks \cite{yang2021superb, mohamed2022self}. A large collection of methods have been used, including clustering and mixture models \cite{varadarajan2008unsupervised}, but recent work tend to focus two main classes of models: masked prediction models and compressive reconstruction models. The models of the first class learn a representation for speech that can best predict a masked portion of the data based on either past samples \cite{oord2018representation,chung2019unsupervised}, or both past and future samples \cite{schneider2019wav2vec,baevski2020wav2vec, hubert,chen2022wavlm}.  Models of the second class are auto-encoders that learn a discrete latent representation that can reconstruct the input data \cite{defossez2022high,zeghidour2021soundstream}. Here, we use as a illustration of our benchmark a simple model from the first class (contrastive predictive coding, CPC, \cite{oord2018representation}).  
\subsection{Domain adaptation}
Robustness to speaker or domain changes in unsupervised speech models is a common problem. In older HMM-GMM ASR systems, common techniques included feature normalization like VTLN or fMMLR that intended to learn a transformation of the input features in order to match the distributions across domains. Since the advent of deep learning, the focus has been to increase the size of the training set in order to cover as many domains as possible, or conditioning the model with speaker representations. In ~\cite{Spk_resampling} authors try to address the imbalanced amount of data between speakers with a resampling strategy. In \cite{ASN}, the authors use adaptation layers, a technique that we will use in this paper. Other domain adaptation techniques such as Teacher-Student learning \cite{teacherstudent} showed promising results but heavily rely on the availability of huge amount of unlabelled or weakly labelled data. 
Adversarial training \cite{SUN201779,Meng_2017} is a common way of performing domain adaptation, this technique uses a discriminator that is trained to infer the domain from the intermediate features of a model. This discriminator creates a loss that reduces the shift between source and target domain features.

\section{The \dataset{} benchmark}
\subsection{Dataset}
\begin{figure}[t!]
    \centering
    \includegraphics[width=0.5\textwidth]{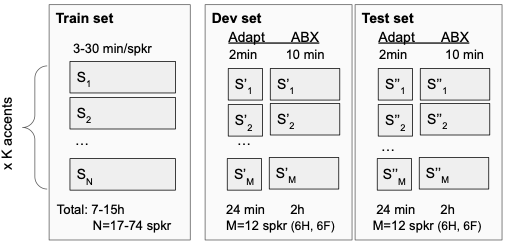}
    \caption{Structure of the \dataset{} benchmkark. Each of the K accent has a training set, a dev and a test set, each containing different speakers (K=10 accents). To allow for speaker adaptation, the dev and test set provide for each speaker a 2min adaptation set. }
    \label{fig:abxdataset}
\end{figure}

We build our benchmark on top of the AESRC dataset~\cite{AESRC} , which contains ten different regional accents. From this dataset we make a split for each accent between a train, a dev and a test set, balancing for male and female speakers (See Figure \ref{fig:abxdataset}).
The Test and dev sets consist of two hours of speech for each accent, with six females and six males (approximately 10 minutes per speaker). In addition, for each of these speakers, we prepare an additional 2:00 minutes for extracting speaker-specific statistics or embeddings for speaker adaptation methods. It allows methods that aim to compensate for the high variability within non native speakers. The rest of the speakers is used in the training set (see Table \ref{tab:ABXtrain} for the duration and number of speakers per accent). 

We use the AESRC transcriptions of the dev and test set which we phonemize (using phonemizer) and force align (using kaldi) to obtain the timestamps of each phonemes. The train set does not have transcriptions.

\begin{table}[t!]
\centering
\caption{Duration and number of speakers for the Train Set of \dataset{}. Prefix "e-" highlights that the data is English speech.}
\begin{tabular}{lccc}
\hline
Accent   & Acronym &Duration& Nb of speakers             \\
\hline
American & e-us & 9:56h &21 H, 25 F \\
British  & e-uk & 15:06h&38 H, 36 F \\
Canadian  & e-ca & 8:07h & 9 H, 10 F  \\
Chinese   & e-ch & 8:23h &14 H, 12 F  \\
Indian   & e-in & 7:34h & 8 H, 10 F  \\
Portuguese & e-pt & 9:18h &13 H, 15 F  \\
Korean  &  e-ko & 8:24h &11 H, 11 F  \\
Japanese & e-ja & 8:25h &11 H, 11 F  \\
Russian  & e-ru & 7:35h & 8 H,  9 F    \\
Spanish  & e-es & 8:06h &10 H, 10 F  \\
\hline

\end{tabular}
\label{tab:ABXtrain}
\end{table}

\begin{table*}[t!]
\center
\captionof{table}{ABX scores (\%) across speakers for CPC models trained on the Old domain (male or female LibriSpeech) and tested on the Old and New domains (the other sex). Lower scores are better and are averaged across males and females experiments. On average the baseline models only trained on the Old domain give 9.06\%  and 11.86\% on the Old and New test sets respectively. Bold correspond to the best scores for each line and underline, overall. FT: simple fine tuning. resamp-FT: fine tuning with resampling. DN: domain normalization. \label{CPC_acc_MF}}
\resizebox{0.8\linewidth}{!}{
\begin{tabular}{l c cc c cc c cc c cc }
\toprule
&& \multicolumn{2}{c}{FT}&& \multicolumn{2}{c}{resamp-FT} && \multicolumn{2}{c}{DN}&& \multicolumn{2}{c}{DN+resamp-FT}\\ \cline{3-4}\cline{6-7}\cline{9-10}\cline{12-13}
{Split \slash Domain} && Old & New  && Old & New        && Old & New          && Old & New     \\ \hline
2min 1 spk     && {9.16} & 11.82    && {9.24} & 11.79   && {\bf 9.06} & 15.22 && {9.10} &\bf 10.8     \\ 
20min 10 spk   && {9.31} & 11.42    && {9.17} & 11.15   && {\bf 9.10} & 13.68 && {9.15} &\bf 9.31     \\ 
2h 60 spk      && {9.41} & 10.62    && {9.05} & 10.02   && {9.22} & 9.76      && {\bf 9.05} &\bf 8.59     \\ 
{8h  240 spk}  && {9.57} & 9.74     && {8.89} & 9.38    && {9.09} & 8.98      && \bf 8.89 &\bf 8.19    \\ 
{16h 479 spk}  && {9.56} & 9.96     && {9.00} & 9.33    && {8.93} & 8.93      &&\underline{\bf 8.34} &\underline{\bf 8.04}     \\ \bottomrule
\end{tabular}}
\end{table*}

\subsection{Metrics}

We provide in our Github repository all the necessary resources to compute the ABX error that have been used in several zero resource challenges~\cite{ZRS2020}~\cite{ZRS2021}. ABX metric evaluate for a pair of sounds representations ($A$, $B$) from for example ("bap", "bop"), the probability that the representation X of another instance of the sound "bap" is closer to $A$ than $B$. ABX error rate is computed by averaging over all the minimal pairs of phone trigrams in the corpus. In this benchmark we focus on the more challenging ABX across-speaker metric, which uses $X$ instances from a different speaker than the one of the pair ($A$,$B$).

\section{Adaptation baseline}

Here, we present a simple baseline system based on Contrastive Predictive Coding (CPC), which illustrates the challenges posed by this benchmark.
\subsection{Contrastive Predicting coding}

	CPC has been introduced in \cite{cpc_oord} as a simple SSL method for learning speech representations \cite{morgane,eugene}. Those models are made of 2 components :
	
	\begin{itemize}
		\item  An encoder that generates a sequence of embeddings ($z_1$, $z_2$, ..) from the audio data.
		\item  An autoregressive network that generates a context $c_t$ from the sequence of embeddings.
	\end{itemize}

The training objective consists in 
	
\begin{equation}
	\mathcal{L}_t = -\dfrac{1}{M} \sum_{m=1}^{M} \log \left( \dfrac	{\exp(z_{t+m}^TW_mc_t)}{\sum_{x\in X} \exp(x^TW_mc_t)} \right)
\end{equation}

Where $W_m$ is a linear classifier, $X$ is the set of negative examples. 

In this work the encoder is made of 5 1-d convolutional layers with 256 channels each. Kernel sizes (10, 8, 4, 4, 4) and strides of (5, 4, 2, 2, 2). The output of every convolutional layer is normalized either using the layer normalization or the adaptive domain normalization that will be discussed in the following section. The context network consists in 2 LSTM layers with 256 hidden dimensions. Similarly to ~\cite{rivire2020unsupervised} the linear function $W_n$ has been replaced with a single layer transformer. In the experiments we use the outputs of the context network as the speech embedding.

As described in ~\cite{rivire2020unsupervised} contrastive loss is particularly effective when the set $X$ of negative examples are taken from the same speaker as the positive one. Using multiple speakers makes the contrastive task solvable only by distinguishing speakers. Using the same speaker for the negative examples forces the model to incorporate speech content into the representations.

\subsection{Adaptive Domain normalization}

Normalizing features tend to make the embeddings more invariant to the speaker and thus, focusing more on the semantic content of the speech. This work applies an adaptive normalization ~\cite{ASN} in various settings to see which kind is most resilient to new speakers. 

Formally with a set of domains $X_1, .., X_D$, denote $h_t^{l-1}$ as the p dimensionnal output of the model's $l-1^{\text{th}}$ layer. In order to reduce the computational cost of the normalization, a non linear transformation is applied such that :

\begin{equation}
	g_t^{l-1} = \tanh(W_g h_t^{l-1} + b_g)
\end{equation}

Where $W_g$ is a $d_g \times p$ weight matrix with $d_g << p$. One can then use a weighted summation of the frames from a domain $d$. Using a softmax weighting on obtain the following set of weights :

\begin{equation}
	\alpha_{d,t} = \dfrac{\exp(\text{mean}(g_{s,t}^{l-1}}{\sum_{\tau} \mathbb{1}[\tau \in X_d]\exp(\text{mean}(g_{s,t}^{l-1}))}
\end{equation}

Then the context vector of the domain d can be computed with :

\begin{equation}
	c_d = \sum_t \alpha_{d,t} \mathbb{1}[\tau \in X_d] g_t^{l-1}
\end{equation}

One can train 2 linear layers $\mathcal{R}^d_g \rightarrow \mathcal{R}^p$ to output the new scales and biases $\gamma_s^l$ and $\beta_s^l$. 

\begin{equation}
\begin{aligned}
    \gamma_s^l &= W_\gamma^l c_d + b_\gamma^l \\
	\beta_s^l &= W_\beta^l c_d + b_\beta^l,
\end{aligned}
\end{equation}

Finally the normalized output of the layer $l-1$ is given by:
\begin{equation}
	\tilde h_{d,t}^l = \hat h_{d,t}^{l-1} \gamma_d^l + \beta_d^l
\end{equation} 

We elaborate more on the different settings for the domain choices in the experiments part (Sec~\ref{sec:Experiments}).

\subsection{Domain Adaptation}

Domain adaptation is a task consisting of taking a model pretrained on domains $\mathcal{D}_0$,~\dots $\mathcal{D}_{N-1}$ that usually contains a lot of data and trying to apply this model to different domains $\mathcal{D}_N$,~\dots $\mathcal{D}_{N + N' -1}$ that usually have a limited amount of available data. In speech processing this task is often related to new speakers who have different accent, or talk in different condition.

Even though retraining the model from scratch on the augmented dataset might be the best performing method, this doesn't allow for fast adaptation to new speakers and is very costly. This work presents a fine tuning approach that aims to quickly adapt to out of domain data with limited computation time and limited data. 

\section{Experiments}
\label{sec:Experiments}

\subsection{Librispeech Male/Female Experiment} \label{sec:MF}

The first experiment conducted to evaluate our methods was to train models on Librispeech considering two domains Male and Female speakers whose labels are available in the metadata. We trained a vanilla CPC model on all female speakers in train-clean to adapt it to male speakers (and conversely). The fine tuning data used consisted of several splits of the female speakers from train-clean, with different amounts of data and speakers. We defined 16 different splits:

\begin{table*}[t!]
\caption{ABX score across speakers within domain, on the accented English test set for different domain adaptation methods (FT: fine tuning with resampling, DN: domain normalization). We boldface the best results per columns.\label{table: ABX_accents}}
\resizebox{\linewidth}{!}{
\begin{tabular}{l c c c c c c c c c c l}
\hline
                 & e-us & e-uk & e-ca & e-ch & e-in & e-ja& e-ko & e-pt & e-ru & e-es & Average \\ \hline
MFCC             & $37.9$    & $37.3$   &$ 35.5 $   & $32.7$   &$ 33.5$  & $36.2$    & $32.6$  & $35.0$      & $35.2$   & $35.6$   & $35.1 \pm  2$ \\ 
AESRC            & $16.1$    & $14.9$   & $22.4$    & $16.0$   & $17.1$  & $20.1$    & $21.9$  & $15.5$      & $18.6$   & $16.3$   & $17.9 \pm 2$  \\ 
LS pretrain         & $13.1$    & $13.2$   & $20.3$    & $14.2$   & $15.4$  & $19.4$    & $20.6$  & $13.7$      & $16.5$   & $14.7$   & $16.1 \pm 3 $ \\ 
\hline
FT (single)        &    $12.9$   & $12.9$          &  $20.6$          & $13.9$          &    $16.0$      &     $10.9$       &     $20.4$     &      $13.6$        & $13.0$ &  $12.7$ & $14.7\pm 3$\\
FT (joint)         & $\bf 12.3 $   & $12.4$   & $20.1$    & $13.4$   & $11.7$  & $10.5$    & $19.9$  & $13.4$      & $12.2$   & $12.1$   & $13.8 \pm 3$  \\ 
DN + FT (single) & $13.0$    &  $13.3$    &      $20.3$    &$12.8$   & $\bf 11.0$  & $9.8$    &   $20.0$   &    $18.2$        & $\bf 9.9$   & $11.7$   &    $14.0 \pm 4$    \\ 
DN + FT (joint)  & $13.2$    & $\bf 11.8$   & $\bf 18.2$    & $\bf 9.0$  & $11.7$  & $\bf 9.7$    & $\bf 13.4$  & $\bf 12.1$  & $12.1$   & $\bf 11.6$  & $\bf 12.3  \pm 2$  \\ 
\hline
\end{tabular}}
\end{table*}

As a primary baseline, the fine tuning is performed using only the data from the new domain, this experiment is referred to as the simple baseline. In all other experiments the data from the original domain is also used at fine tuning time. However, the new domain is oversampled so that half of the batches come from a male speaker and the other half from a female speaker. This helps regularize and avoid catastrophic inferences.

The average CPC accuracy is the validation metric used for the early stopping. Empirical experiments show that it is correlated with the ABX score. 
As can be seen in the simple baseline, the performance on the original domain tend to drop significantly while training on the new one. This leads to very early training stop if there is not enough data to make the increased performance on the new domain to compensate for the drop on the original domain. On the contrary using the original domain during fine tuning with even sampling keeps the performance on females very stable which helps the model perform more epochs while increasing the average accuracy. This experiment shows the benefits of keeping data from the original domain when adapting to a different one. From this observation, all the later experiment use this resampling strategy.
The last two columns describe the effect of the domain normalization. The new domain statistics stored into the domain normalization layer are initialized with the statistics of the original domain. With this normalization, fine tuning all the weights causes the model to drop to drop significantly in performance in the first steps losing the benefits of the pretraining and resulting in very slow convergence comparable to retraining from scratch. To deal with this issue, we included a warm up fine tuning stage were the weights of the network are frozen, only speaker statistic and weights $W_g, W_{\beta}, W_{\gamma}$ are updated. Results from both ABX tests demonstrate that more data is needed to benefit from only updating the domain normalization. 
The last column describes the performance of the model after continuing training of previous models with weights unfrozen. As can be observed, this fine tuning framework outperforms other methods in every case. We even report increased performance while using 8x less data for the new domain. One limitation of this approach is that it requires having the domain label as input. In the Male/Female case, this is not an issue since gender is known to be easily obtainable (e.g. with a linear classifier over MFCC features). However this issue is much more relevant in the case of accented speech when unaware of the utterance's domain. 

\subsection{\dataset{} Adaptation experiments}

This section presents domain adaptation experiments on \dataset{}. We used the best methods from section \ref{sec:MF} and apply them in our benchmark. We considered Librispeech clean to be the original domain on which models are pretrained. The models are then fine tuned with the accented speech data. During this fine tuning phase Librispeech samples are sampled according to our resampling strategy. We provide on our github page all necessary material to construct the same splits used in our experiments. 

Table \ref{table: ABX_accents} sums up ABX scores of the benchmark data over some of the accents from \dataset{}, ABX across refers to ABX calculated across speakers from the same accent. AESRC Training refers to straightforward CPC training on \dataset{} without initial pretraining, using pretraining on Librispeech significantly boost performances even in the simplest experiments. Experiments referred as "joint" are performed over all the accents i.e. there is a single model per line. On the contrary other line report results from different model for each accent. 
The resampling strategy was used in all the fine tuning experiments. In the single domain finetuning experiments, sample alternate between librispeech samples and samples from the accent. In the "joint" experiments we sample in turn from the eleven domain (ten accents and librispeech).

Similarly to the preliminary experiments the domain normalization warm up followed by complete fine tuning tends to have the best performances. In average we found that models specialised on a specific domain achieve better results. However in some cases we report that there can be benefits to have more data even from different domains. It can be observed that using all the data available overall improve the model performances on the several accents. Even though some for some accents (e.g. Russian) the specialized model fine tuned only with one domain achieves better ABX score than the model trained on the complete training set.

\section{Conclusion}
We presented a new accent adaptation benchmark for self-supervised speech representation learning which we tested on baseline systems based on an adaptive normalization method. We demonstrated with a toy male/female domain split on librispeech that this method has the potential of improving ABX scores by about 33\% relative. Applied to the accent dataset, it show improvements of 23\% relative, suggesting that the accent adaptation problem is more difficult and that there is room for improvement. We hope that this benchmark will trigger further work in domain adaptation. 

\section*{Acknowledgments}
This work was performed using HPC resources from GENCI-IDRIS (Grant 2024-AD011014739R1) and was supported in part by the Agence Nationale pour la Recherche (ANR-17-EURE-0017 Frontcog, ANR10-IDEX-0001-02 PSL*). ED and MK in their EHESS roles were funded by an ERC grant (InfantSimulator). Views and opinions expressed are those of the authors only and do not necessarily reflect those of the European Union or the European Research Council. Neither the European Union nor the granting authority can be held responsible for them.
\bibliographystyle{IEEEtran}
\bibliography{refs}

\end{document}